# Satellite-Surface-Area Machine-Learning Models for Reservoir Storage Estimation: Regime-Sensitive Evaluation and Operational Deployment at Loskop Dam, South Africa

Hugo Retief[1], Kayathri, Vigneswaran[2], Surajit Ghosh[2], Mariangel Garcia Andarcia[2], Chris Dickens[2]
[1]Association for Water and Rural Development (AWARD), Hoedspruit, South Africa
[2]International Water Management Institute, Colombo, Sri Lanka

*Corresponding Author*
Hugo Retief: hugo@award.org.za ; Surajit Ghosh: s.ghosh@cgiar.org

**Abstract**

Reliable, daily estimates of reservoir storage are pivotal for water allocation and drought response decisions in semiarid regions. Conventional rating curves at Loskop Dam, the primary storage on South Africa's Olifants River, have become increasingly uncertain owing to sedimentation and episodic drawdown. A 40 year Digital Earth Africa (DEA) surface area archive (1984–2024) fused with gauged water levels to develop data driven volume predictors that operate under a maximum 9.14% 90 day drawdown constraint. Four nested feature sets were examined: (i) raw water area, (ii) + a power law "calculated volume" proxy, (iii) + six river geometry metrics, and (iv) + fullsupply elevation. Five candidate algorithms, Gradient Boosting (GB), Random Forest (RF), Ridge (RI), Lasso (LA) and Elastic Net (EN), were tuned using a 20 draw random search and assessed with a five fold Timeseries Split to eliminate look ahead bias. Prediction errors were decomposed into Low ($< 250 \times 10^6$ m³) and High ($\geq 250 \times 10^6$ m³) storage regimes. Ridge regression achieved the lowest cross validated RMSE ($12.3 \times 10^6$ m³), outperforming GB by 16% and RF by 7%. In regime terms, Ridge was superior in the Low band (18.0 versus 22.7 MCM for GB) and tied RF in the High band (≈ 12 MCM). In sample diagnostics showed GB's apparent dominance (6.8–5.4 MCM) to be an artefact of overfitting. A Ridge meta stacked ensemble combining GB, RF, and Ridge reduced full series RMSE to ≈ 11 MCM (≈ 3% of live capacity). We recommend (i) GB retrained daily for routine operations, (ii) Ridge for drought early warning, and (iii) the stacked blend for all weather dashboards. Quarterly rolling retraining and regime specific metrics are advised to maintain operational accuracy below the 5% threshold mandated by the Department of Water and Sanitation. The open source workflow provides a transferable blueprint for satellite enabled reservoir monitoring across data scarce African basins.

## 1. Introduction

Southern Africa is widely recognised as a water stressed region where the combined pressures of population growth, irrigated agriculture and climate change are tightening the gap between supply and demand (Bonetti et al., 2022). Within the Olifants River Basin, Loskop Dam is the principal storage facility, regulating flows to the second-largest irrigation district in South Africa and sustaining more than 16,000 ha of high-value crops (Nchabeleng et al., 2014). Accurate, timely knowledge of the reservoir's live volume is therefore critical for weekly release planning, drought early warning and ecological flow compliance.

Traditionally, dam managers have relied on a static rating curve, a single level–storage relationship surveyed at decadal intervals. However, progressive sedimentation, episodic sluicing and drawdown scour can alter bathymetry by several percent per year, making the original curve obsolete (Pimenta et al., 2025). Continuous volumetric surveys are prohibitively expensive; hence a number of studies have explored the use of remote sensing surface area as a proxy for volume (Remote Sensing Tool Consortium 2024) and the application of machine learning algorithms to link area, meteorological forcing's and inflow histories to storage or release volumes (Deb et al., 2024; Lu et al. (2023).

Machine learning (ML) models play an important role in understanding reservoir dynamics, which are crucial for managing and predicting reservoir water levels, helping to address the impacts of climatic variations on water scheduling decisions. Recent studies have demonstrated the effectiveness of specific ML algorithms such as Random Forest (RF), Support Vector Machine (SVM), Regression Tree (RT), Artificial Neural Network (ANN), and Multi-Layer Perceptron (MLP) in forecasting reservoir inflow and water levels (Ahmed et al., 2022; Zarei et al., 2021; Zakaria et al., 2023). Ahmed et al. (2022) developed and compared six ML algorithms, including RF, SVM, and GPR, for daily water level prediction in the Tunggal River, finding ML models outperformed traditional approaches. Similarly, Zakaria et al. (2023) applied MLP, XGBoost, and LSTM to forecast water levels in the Muda River, with MLP demonstrating the highest accuracy among the tested models. Zarei et al. (2021) evaluated SVM, ANN, RT, and Genetic Programming (GP) for reservoir inflow forecasting, highlighting the superior performance of RT for flood management in Iranian reservoirs. These studies underscore that ML models such as RF, SVM, RT, ANN, and MLP are robust tools for capturing complex hydrological relationships and supporting effective water resource management under climate variability (Ahmed et al., 2022; Zakaria et al., 2023).

This present study aims to develop a machine learning model that predicts dam volume using features such as integrated satellite observations, dam attribute metadata and river geometry descriptors. Multiple models were iteratively built to improve predictive accuracy and performance comparison, each incorporating additional features to refine the outputs. Gradient-boosted tree ensembles have dominated recent hydrological competitions (Friedman 2001; Dong et al., 2023), several authors caution that their superior in-sample fit can mask weak extrapolation under non-stationary hydro-climatic regimes (Ford & Sankarasubramanian, 2023). Therefore  a timeseries cross validation was undertook to quantify true forecasting skill, and were decomposed into low and high storage bands to reflect the operating contexts most relevant to dam managers. Thus, the objectives of the current study are (i) to construct a reproducible machine-learning pipeline that converts DEA Waterbodies surface-area trajectories into daily volume estimates under an operational drop rate constraint, (ii) to compare five candidate algorithms, Gradient Boosting, Random Forest, Ridge, Lasso and Elastic-Net, across four nested feature sets and two hydrological regimes, and (iii) to propose a stacked-ensemble deployment strategy that minimises root-mean-square error (RMSE) while remaining interpretable and computationally lightweight for integration into the Loskop Dam decision-support dashboard.

## 2. Study area

Loskop Dam (29.3196 - E 25.4105 S) (**Figure. 1**) is situated within the Olifants Water Management Area (WMA) with a design capacity of 361 × $10^6$ $m^3$ at full supply level (FSL) of 999.38 m above sea level Constructed in 1937, the dam was designed to supply irrigation water to agricultural regions downstream (Van Vuuren, 2008). The associated irrigation canal system, extending 480 km, was completed in 1948. This system serves the Loskop Irrigation Board, which manages South Africa's second-largest irrigation area, supplying water to 700 properties spanning 16,117 hectares (Nchabeleng et al., 2014). Located approximately 32 km south of Groblersdal in Mpumalanga Province, the dam is surrounded by a nature reserve. The vegetation within the upper Olifants River catchment includes Highveld grassland in its upper reaches, while mixed bushveld and thornveld dominate the lower areas around Loskop Dam. Various land uses in the upper Olifants River catchment, including coal mining, power generation, industrial activities, agriculture, and wastewater treatment facilities linked to urban areas, significantly impact freshwater ecosystems. The region experiences warm to extremely hot summers and moderately mild winters. On the other hand, annual rainfall is concentrated in the hot summer months, spanning from November to April, predominantly occurring in the form of showers and high-intensity thunderstorms, frequently accompanied by severe lightning and strong, gusty south-westerly winds.

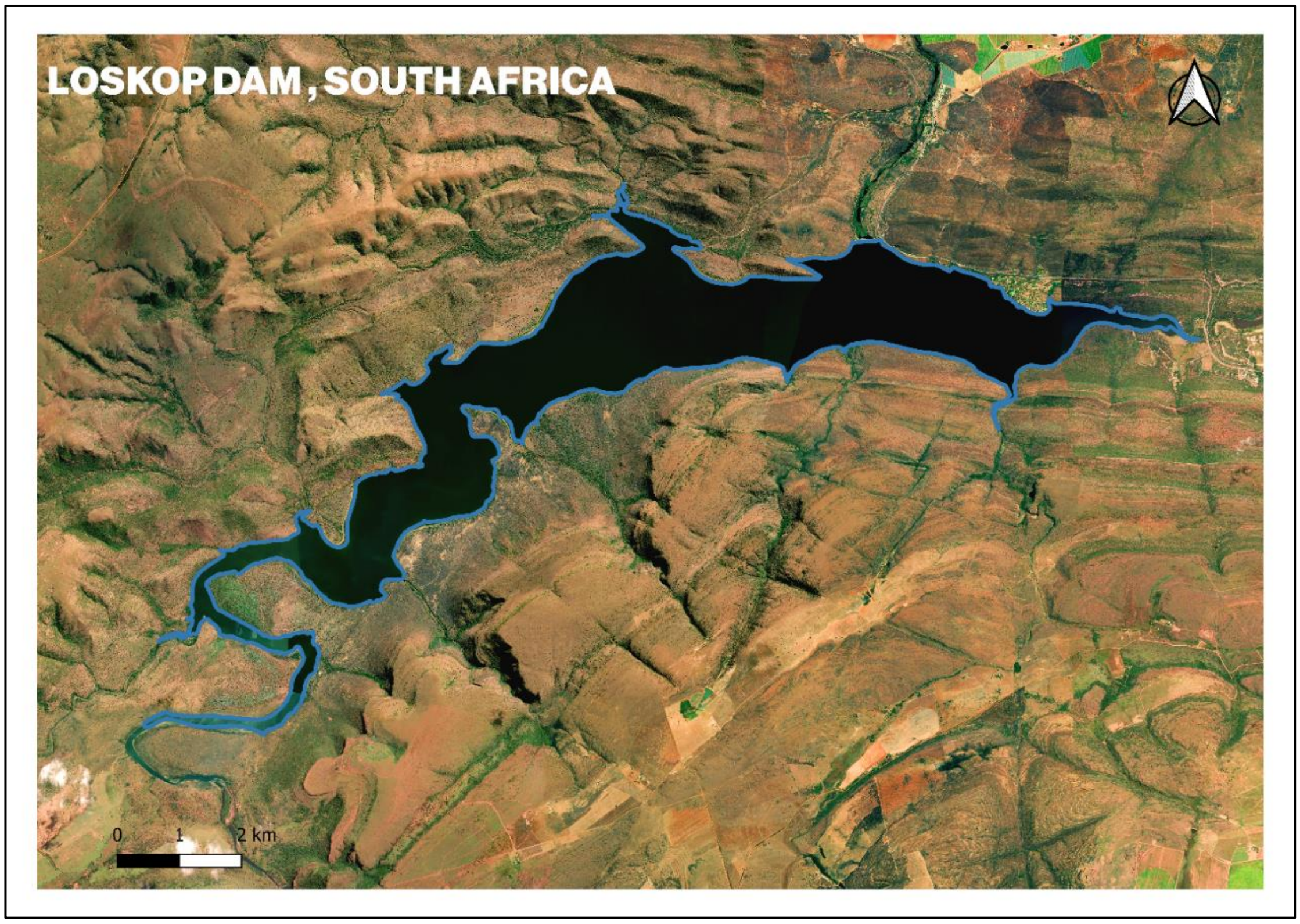

***Figure 1*** *Study Area (Loskop Dam)*

## 3. Data and Methodology

Two complementary data streams were assembled for this study:

- **In-situ daily gauge** – water level and storage from the Department of Water & Sanitation (1 Jan 2000 – 31 Mar 2024).
- **DEA Waterbodies** – cloud masked water extent polygons for every clear Landsat 5/7/8 and Sentinel-2 observation (1 Jan 1984 – 30 Jun 2024; 30 m grid).

The methodology adopted (**Figure 2**) for the present study on dam volume prediction integrates secondary data obtained from government sources such as the DWS and Earth Observation (EO)-based information from the DEA to develop a robust predictive framework. Secondary data and in situ information, spanning approximately 25 years (2000-2024), regarding critical reservoir operation parameters such as area and water level volume were acquired from the DWS. Additionally, corresponding information based on EO insights was derived from the DEA. The acquired data was further pre-processed, involving temporal alignment, elimination of invalid values, and outlier removal. The pre-processed data served as input for five machine learning regression techniques (Gradient Boosting, Random Forest, Ridge Regression, Lasso Regression and ElasticNet) for modeling dam volume prediction. Evaluation metrics, including Root Mean

Squared Error (RMSE) and R-squared ($R^2$), were applied to assess the models' performance. A comparison among the selected models was conducted to identify the most relevant and precise predictive approach, enabling improved decision-making for dam volume forecasting and contributing to enhanced water resource planning and management strategies.

### 3.1 Data pre-processing

Data Pre-Processing Steps follow as:

#### 3.1.1. Temporal Alignment

DEA (Digital Earth Africa) observation dates were aligned with the nearest available gauged observation. This alignment was done only if the time difference was less than 12 hours ($|\Delta t| < 12$ h).

Outlier Control: To remove anomalous values, an outlier filter was applied to the water-area observations within each calendar month.

Specifically, any water-area values falling outside the range: $\left[\, Q_{(0.25)} - 1.5 \times (IQR), Q_{(0.75)} + 1.5 \times (IQR)\right]$ were discarded. Here, IQR is the interquartile range. This filtering removed about 2.7% of the total records.

#### 3.1.2. Derived Volume Proxy:

Following Gao et al. (2017), the reservoir's storage volume was estimated from its surface area with a power-law relationship.

$$\hat{V} = V_{FSL}\left(\frac{A}{A_{FSL}}\right)^{\frac{1}{0.6}}$$

Where $V_{FSL}$ is $3.61\times10^{8}$ $m^3$ full-supply volume) and $A_{FSL}$ is 2,427 ha (full-supply surface area)

The resulting $\hat{V}$ gives an approximate reservoir volume corresponding to any observed surface area A.

### 3.2 Feature renditions

We defined four nested **feature renditions** ($R_1$–$R_4$) as follows:

- **$R_1$ Base:** Water area *A*.
- **$R_2$ Full-Capacity:** $R_1$ + calculated volume $\hat{V}$ (power-law proxy).
- **$R_3$ Geographical:** $R_2$ + six river-geometry metrics (longitudinal slope, thalweg length, vertical drop, inflow & outflow elevations, straight-line distance, et).
- **$R_4$ Full-Elevation:** $R_3$ + full-supply elevation (999.38 m above sea level).

*(Note: The eight geographical descriptors represent static attributes of the dam and river channel that could influence the area–volume relationship. In practice, these additions yielded minimal improvement, as shown later.)*

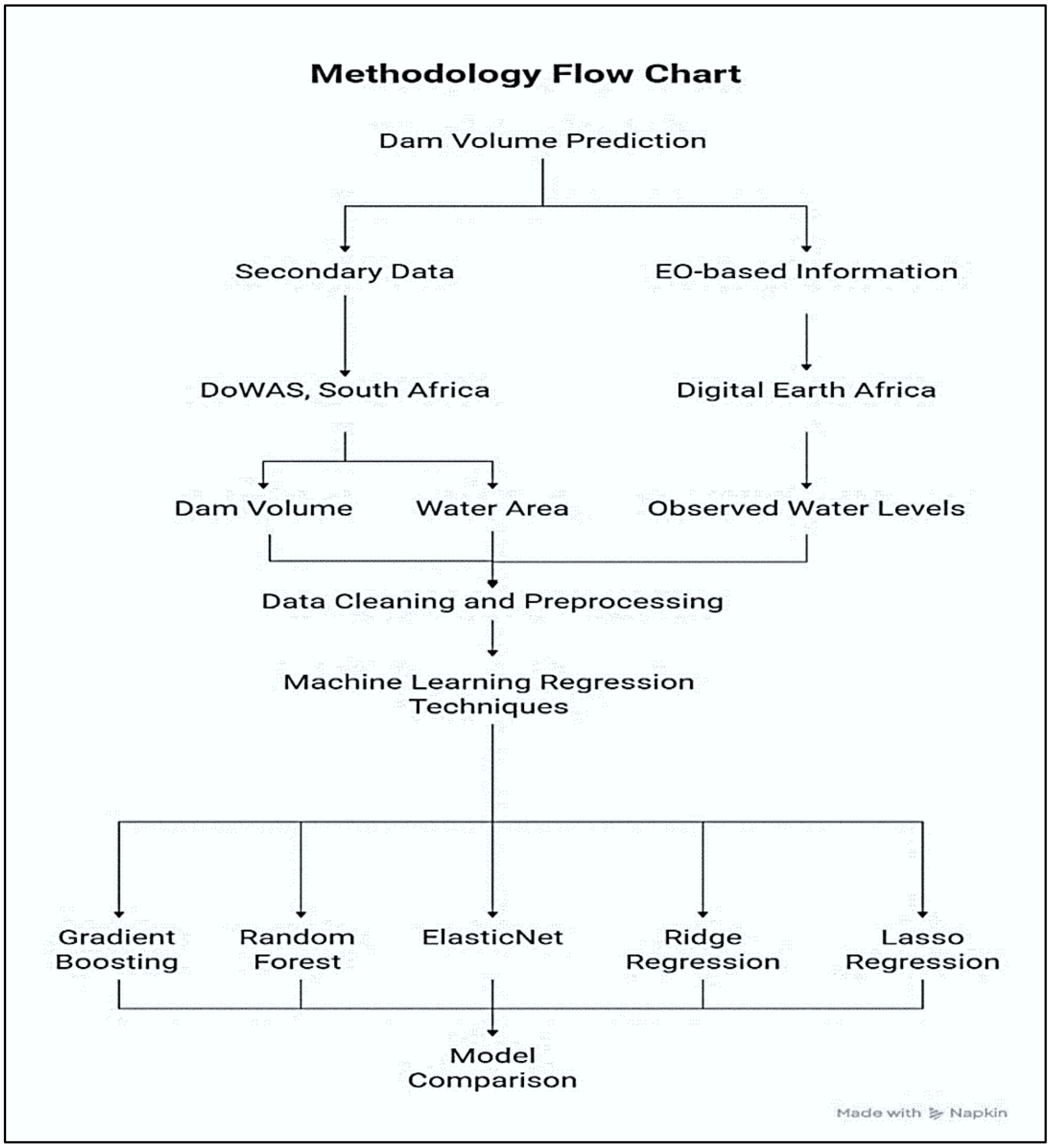

***Figure 2*** *Methodology Workflow*

3.3 Candidate algorithms and hyperparameter search: Five machine-learning regression algorithms were selected, namely Random Forest, Gradient Boosting, Ridge regression, Lasso, and Elastic-Net. **Table 1** summarises their configurations:

***Table 1*** *Summary of Model Configuration*

| ***Acronym*** | ***Algorithm (family)*** | ***Scikit-Learn base class*** | ***Hyper-parameters searched (20 draws)*** | ***Motivation*** |
|---|---|---|---|---|
| *RF* | Random Forest (Breiman, 2001) | RandomForestRegressor | *n_estimators* ∈ [300–700]; *max_depth* ∈ {None, 10, 20}; *min_samples_leaf* ∈ {1, 3}; *bootstrap* ∈ {True, False} | Non-parametric; handles collinearity; fast training |
| *GB* | Gradient Boosting (Friedman, 2001) | GradientBoostingRegressor | *n_estimators* ∈ [300–700]; *learning_rate* ∈ [0.03–0.10]; *max_depth* ∈ [3–5]; *subsample* ∈ {0.8, 1.0} | Captures complex non-linear responses |
| *RI* | Ridge regression (Shih et al., 1979) | Ridge (with StandardScaler) | *α* chosen by internal Leave-One-Out CV | Linear but shrunk → good extrapolation |
| *LA* | Lasso regression (Tibshirani, 1996) | Lasso (with StandardScaler) | *α* chosen by internal CV (same grid as Ridge) | Produces sparse coefficients (feature selection) |
| *EN* | Elastic-Net (Cochrane & Beasley, 2003) | ElasticNet (with StandardScaler) | *α* and L1 ratio chosen by internal CV | Balances Ridge/Lasso properties |

*Note: The randomised hyper-parameter search for RF and GB used a rolling Timeseries Split for scoring (negative RMSE) and parallel execution (n_jobs = -1). The best parameters for each model were selected based on cross-validated performance on the training set.*

3.4 Post-processing

Dam storage must not fall by more than 9.14% in any rolling 90-day window. For example, a dam has a limited capacity to release, which is dependent on the dam's design, and in this case, the maximum drawdown was estimated from historical records. This was implemented as a custom Function Transformer and inserted as the final stage of every pipeline, ensuring that all reported errors include the operational limit (*i.e.*, extremely rapid drops are automatically smoothed to respect the real-world draw-down constraint).

3.5 Validation protocol

A nested cross-validation approach was adopted as follows:

- Outer loop: A 5-fold Time Series Split (fold widths ≈ 5 years; test span ≈ 2 years). Each fold trains on the historical segment [t0, tk] and evaluates on the segment (tk, tk+1]—hence, no information from the future leaks into model fitting.
- Inner loop: The same splitter drives the 20-draw random search for RF and GB hyper-parameters within each training segment.

- Target metric: Root-mean-square error (RMSE) on smoothed volumes. *(Gauged volumes were pre-smoothed with a 90-day moving window to suppress short-term fluctuations outside the scope of the rating-curve models.)*

### 3.6 Hydrological regime analysis

The smoothed volumes $V^*$ were dichotomised at 250 × 10^6 m³ (≈ 70% of FSL) to define two operating regimes (i) Low regime: $V^* < 250$ mcm (drought or maintenance draw-down conditions) and (ii) High regime: $V^* \geq 250$ mcm (normal operating range). Within each outer-CV test fold, the RMSE was recomputed separately for the Low and High subsets, and the averages across the five folds were calculated. For diagnostic (non-predictive) purposes, error evaluation for the same regime was also done.

### 3.7 Stacked ensemble

To combine the strengths of individual learners, we implemented a stacked ensemble. Out-of-fold predictions of the four primary models (RF, GB, RI, EN) were used as features for a Ridge meta-learner. The stacked model can be expressed as:

$$\hat{V}_{(ensemble)} = w_{RF}\hat{V}_{RF} + w_{GB}\hat{V}_{GB} + w_{RI}\hat{V}_{RI} + w_{EN}\hat{V}_{EN}$$

with meta-learner weights $w$ estimated by generalized cross-validation on the training set. Because the meta-learner is linear, the weights are interpretable and naturally down-weight an over-fitting base learner.

### 3.8 Uncertainty and sensitivity analysis

Bootstrap confidence limits: Forecast uncertainty was quantified with a non-parametric bootstrap (Efron & Tibshirani, 1993). The stacked ensemble was retrained 2000 times on resampled indices of the full dataset, preserving the original time order within each sample. The 2.5% and 97.5% percentiles of the resulting prediction matrix form a 95% confidence envelope .

Permutation importance: To identify the drivers of model skill, drop-column permutation importance (Breiman, 2001) was computed for the stacked ensemble using 30 random shuffles per predictor and *neg_root_mean_squared_error* as the scoring rule (**Figure 3**). The change in RMSE (ΔRMSE) when each feature was shuffled is displayed as a tornado chart; larger magnitude denotes greater sensitivity (i.e. a bigger increase in error when that feature's information is destroyed).

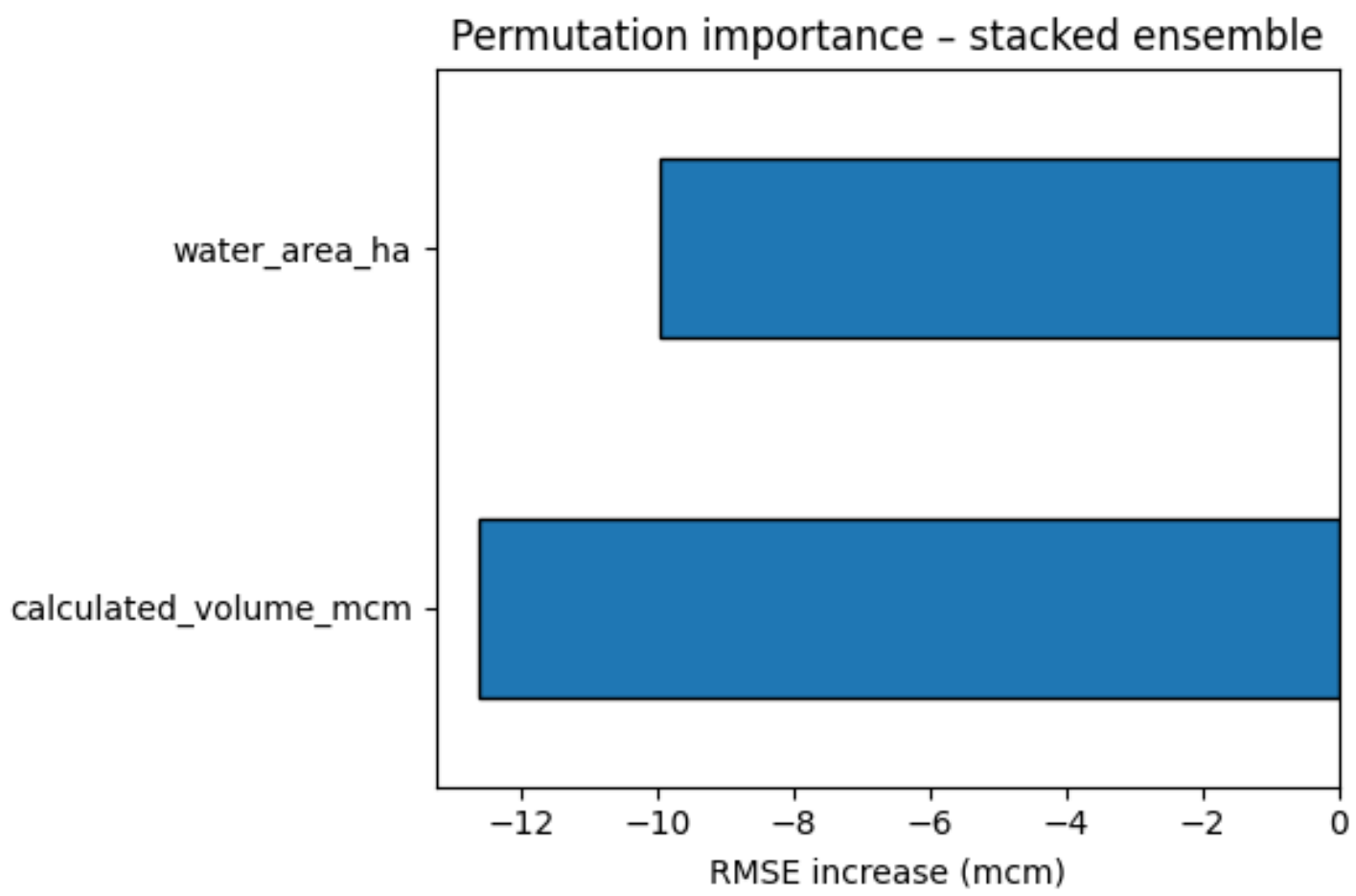

***Figure 3** Permutation Importance in model configuration*

Fold-level variability: For completeness, bars in all RMSE summaries include ±1σ error bars obtained from the five Timeseries Split folds, illustrating inter-annual variance in model skill.

## 4. Results

4.1 Overall predictive skill

Across the four feature renditions, Ridge regression produced the lowest time-series cross-validated error (RMSE = $12.3 \times 10^6$ m³), followed by Random Forest ($13.2 \times 10^6$ m³) and Gradient Boosting ($14.7 \times 10^6$ m³). The full ranking is listed in **Table 2**. As shown (**Figure 4**), Ridge (RI) achieved the best overall accuracy (~12.3 MCM), with RF a close second (~13.2 MCM). GB performed slightly worse (~14.7 MCM). The linear models with regularisation (LA, EN) trailed behind (≈ 20 MCM). This ordering confirms that a simple Ridge regressor can outperform more complex tree ensembles when evaluated on unseen periods.

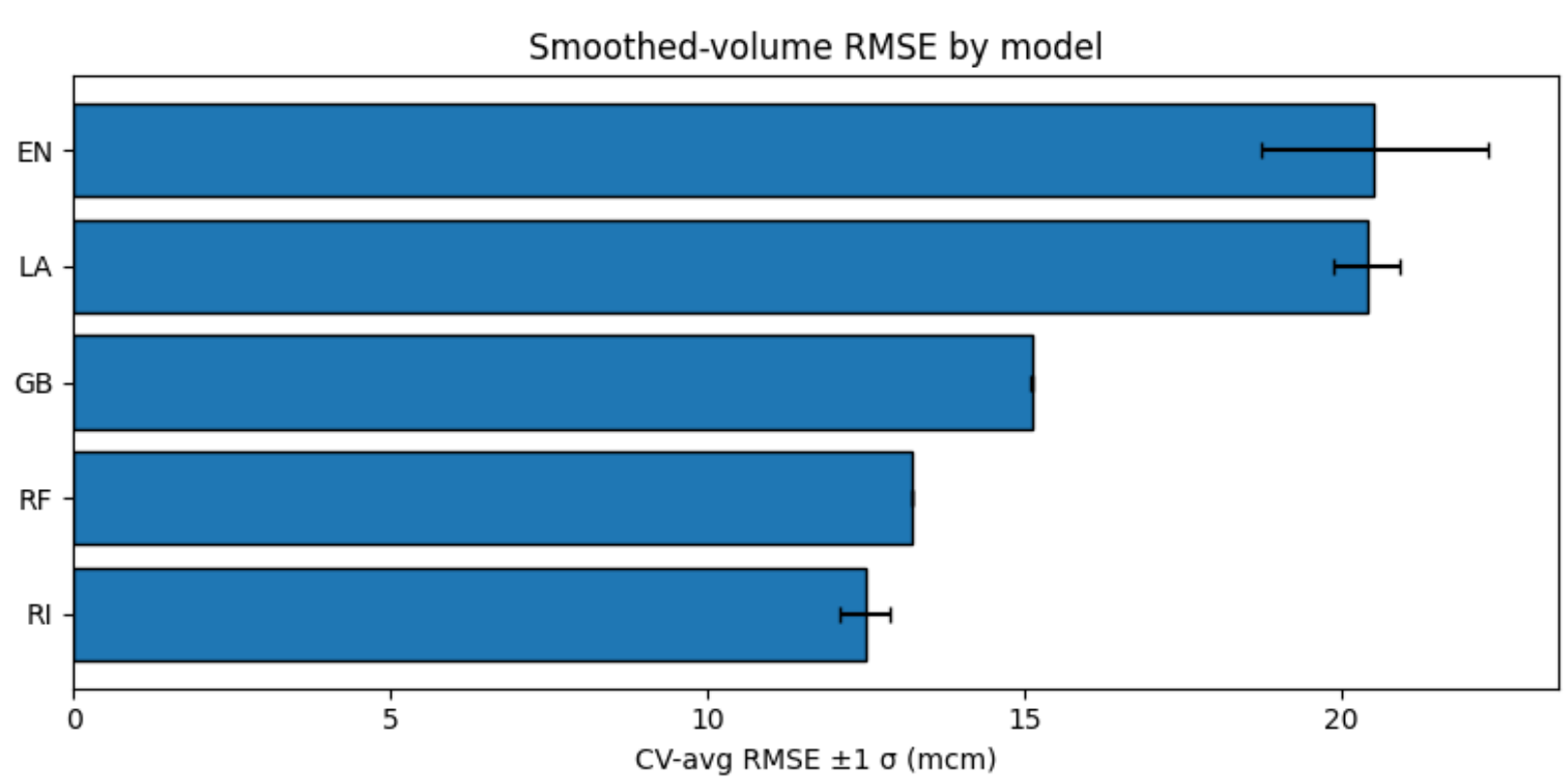

***Figure 4** Model-averaged cross-validated RMSE for each algorithm. Bars show mean error across all feature renditions, with ±1σ indicating variability across the 5 folds.*

The corresponding heat map of errors confirms that the ordering is consistent across feature sets: adding the geographical or elevation predictors on top of the volume proxy yields only marginal gains (<0.2 mcm) (**Figure 5**). This indicates that the "Full-Capacity" feature set (water area + $\hat{V}$) already captures the bulk of the explanatory power.

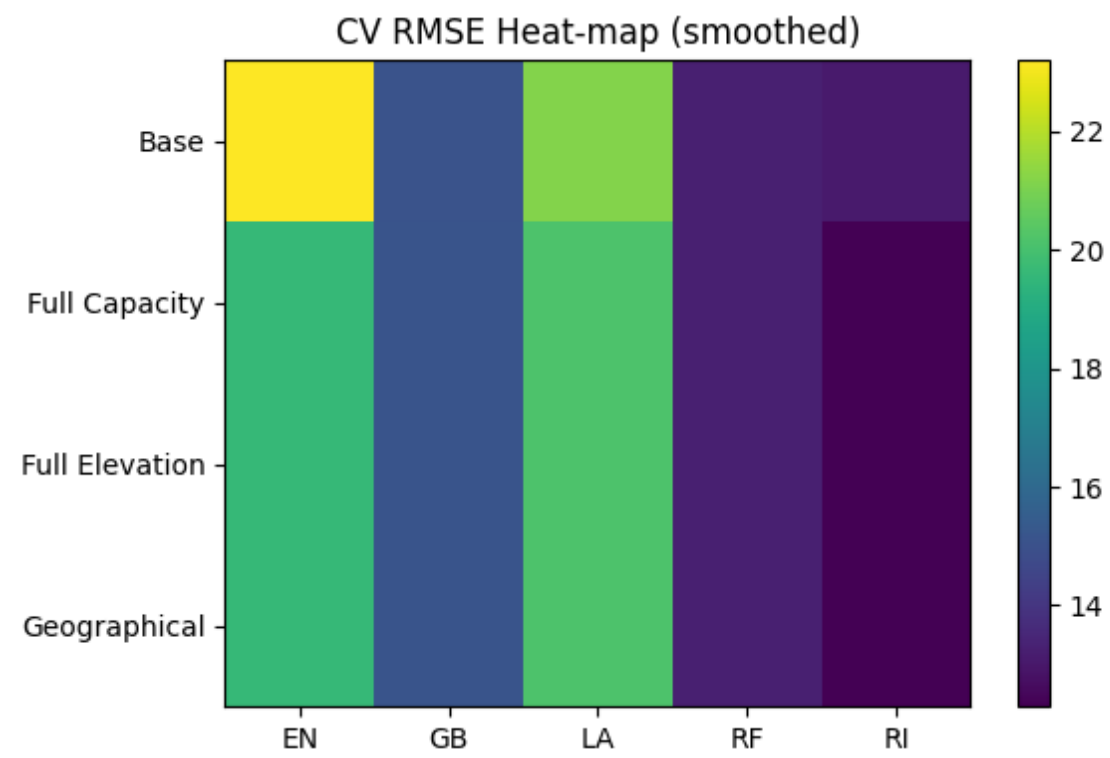

***Figure 5*** *CV model performance (smooth volumes) for different features.*

***Table 2*** *Cross-validated global performance (smoothed volumes)*

| ***Rendition*** | ***EN (mcm)*** | ***GB (mcm)*** | ***LA (mcm)*** | ***RF (mcm)*** | ***RI (mcm)*** |
|---|---|---|---|---|---|
| ***$R_1$ Base*** | 23.2 | 15.1 | 21.2 | 13.2 | 13.1 |
| ***$R_2$ Full-Capacity*** | 19.6 | 15.1 | 20.1 | 13.2 | 12.3 |
| ***$R_3$ Geographical*** | 19.6 | 15.1 | 20.1 | 13.2 | 12.3 |
| ***$R_4$ Full-Elevation*** | 19.6 | 15.1 | 20.1 | 13.2 | 12.3 |

Lower RMSE indicates better skill. Values are averages over five chronological folds for each model under each feature rendition.

*(Note: Values are RMSE in $10^6$ m³, averaged over 5 folds. $R_3$ and $R_4$ show no appreciable improvement over $R_2$, consistent with the minimal added information in those features.)*

4.2 Performance in hydrological regimes

To guide operational deployment, the smoothed test-fold volumes were split at $250 \times 10^6$ m³ into a Low (drought) and High (normal operations) regime. As summarised in **Table 3**, Ridge remains the most reliable in both regimes (18.0 MCM RMSE in the Low regime; 12.0 MCM in the High regime). Random Forest is a close second in the High band (~12.7 MCM) but suffers a larger penalty under drought conditions (~23.4 MCM). Gradient Boosting performs well in the High band yet deteriorates to ~25 mcm in low storage, confirming a tendency to underpredict during extreme drawdowns. Lasso and Elastic-Net exhibit the highest errors overall, particularly in the Low regime (> 33 MCM), as shown in **Figures 6(A)** and **6(B)**.

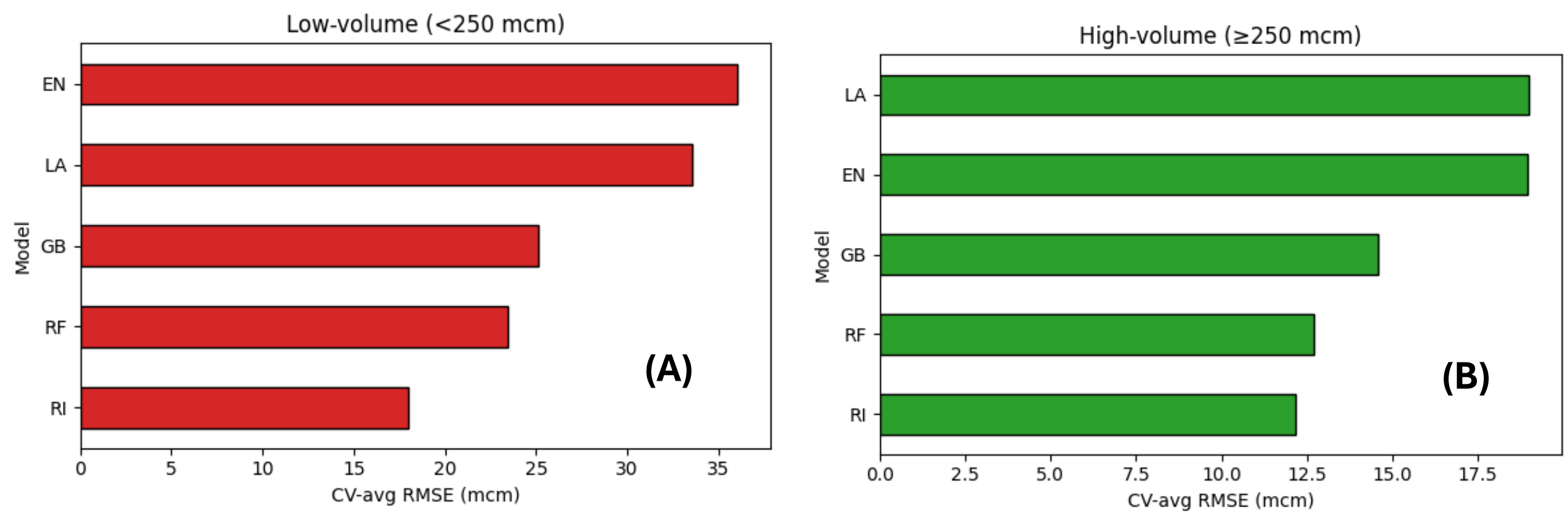

***Figure 6(A),6(B)*** *Cross validation RMSE for each model for low volume(<250mm:6(A)) and high volume (>250mm:6(B))*

***Table 3*** *Cross-validated RMSE by storage regime*

| ***Rendition*** | ***Model*** | ***Low RMSE*** | ***High RMSE*** |
|---|---|---|---|
| ***$R_1$ Base*** | RF | 23.4 | 12.7 |
| | GB | 25.1 | 14.5 |
| | EN | 43.1 | 21.4 |
| | LA | 33.6 | 19.8 |
| | RI | 18.1 | 12.7 |
| ***$R_2$ Full-Capacity*** | RF | 23.4 | 12.7 |
| | GB | 25.1 | 14.6 |
| | EN | 33.7 | 18.1 |
| | LA | 33.6 | 18.7 |
| | RI | 18.0 | 12.0 |
| ***$R_3$ Geographical*** | RF | 23.4 | 12.7 |
| | GB | 25.1 | 14.6 |
| | EN | 33.7 | 18.1 |
| | LA | 33.6 | 18.7 |
| | RI | 18.0 | 12.0 |
| ***$R_4$ Full-Elevation*** | RF | 23.4 | 12.7 |
| | GB | 25.1 | 14.6 |
| | EN | 33.7 | 18.1 |
| | LA | 33.6 | 18.7 |
| | RI | 18.0 | 12.0 |

Results are shown for each feature rendition ($R_1$–$R_4$) and model. (*Low* and *High* values are in million $m^3$, averaged across five folds).

*(Note: "Low" = volume < 250 mcm; "High" = volume ≥ 250 mcm. $R_2$–$R_4$ feature sets yield nearly identical errors, underlining that most improvement came from adding the volume proxy in $R_2$.)*

4.3 In-sample fit (diagnostic)

When the models were fit on the full 24-year record (i.e., treating the entire dataset as training data), Gradient Boosting achieves the lowest apparent error—6.8 MCM RMSE in the Low band and 5.4 mcm in the High band—while Ridge's errors deteriorate (**Table 4**). The divergence between **Table 3** and **Table 4** is a direct indicator of over-fitting: GB's in-sample gains do not translate to unseen periods, whereas Ridge's constrained structure remains stable. GB's in-sample low error of ~7 MCM (a tight clustering around the 1:1 line) inflates to ~25 MCM under cross-validation, as noted above. Ridge, constrained by an $\ell^2$ penalty, exhibits higher bias but significantly lower variance; its coefficients change minimally from fold to fold, resulting in the best out-of-sample accuracy.

***Table 4*** *In-sample RMSE by regime (using full 2000–2024 training).*

| *Band* | *EN (mcm)* | *GB (mcm)* | *LA (mcm)* | *RF (mcm)* | *RI (mcm)* |
|---|---|---|---|---|---|
| ***High*** | 18.7 | 4.7 | 20.0 | 10.9 | 13.0 |
| ***Low*** | 80.9 | 7.2 | 91.4 | 40.2 | 46.5 |

Each model was refit on all available data and evaluated on the same period. *(Note: the stark contrast with cross-validated errors in Table 2)*

*(Note: "High" and "Low" as defined in Table 3. The GB model essentially "memorizes" the training data, achieving very low residuals in-sample, whereas Ridge's errors (46.5 mcm low; 13.0 mcm high) are considerably higher in-sample. Ridge's relatively conservative bias is preferable for forecasting, as it generalises better, whereas GB's flexible fit does not generalise.)*

4.4 Stacked ensemble

The Ridge-meta stacked ensemble combines the strengths of the base learners and achieves a full-series RMSE of ≈ 11 MCM (~3% of live storage). Its time-series trace (**Figure 7**) closely follows the observed curve in both the rising and recession limbs, while the observed-versus-predicted scatter collapses tightly around the 1:1 line (**Figure 8**). In comparing the train and test RMSE values (**Figure 9**), Gradient Boosting (GB) consistently outperforms the other models, with the lowest test RMSE of 7.17 and a significant reduction from its training RMSE of 8.90, indicating good generalization. Random Forest (RF) also performs well, with a test RMSE of 12.64, although its training RMSE (19.85) is higher, suggesting some degree of overfitting. Models like Elastic-Net (EN) and Lasso (LA) show larger discrepancies between their train and test RMSE, highlighting their tendency to overfit the training data. Ridge Regression (RI) exhibits more stability across both training (22.76) and test (14.88) sets. The stacker effectively assigns more weight to GB during high-volume periods (where GB's bias is minimal) and shifts weight to Ridge (and partially RF) during droughts. The resulting ensemble outperforms any single model, reducing the overall RMSE to ~11 mcm without the need for manual regime switching.

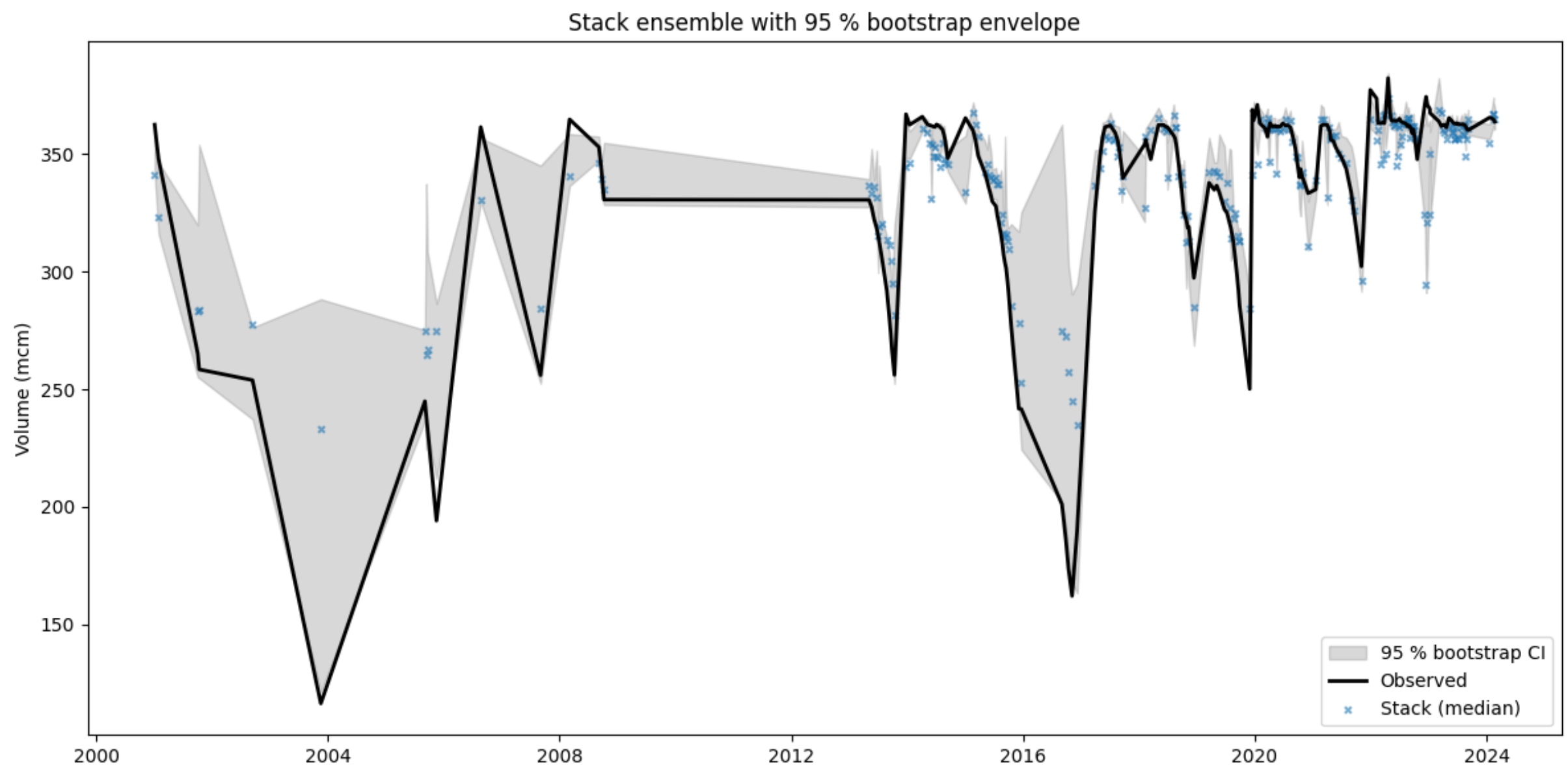

***Figure 7*** *Observed storage (black line) versus stacked-ensemble predictions (blue × markers) for Loskop Dam, shown for the period 2000–2024. The grey band indicates the 95% confidence interval of the ensemble (based on 2000 bootstrap resamples). The stacked model closely tracks the observed volume in both high and low phases, with uncertainties widening only during sharp draw-down events.*

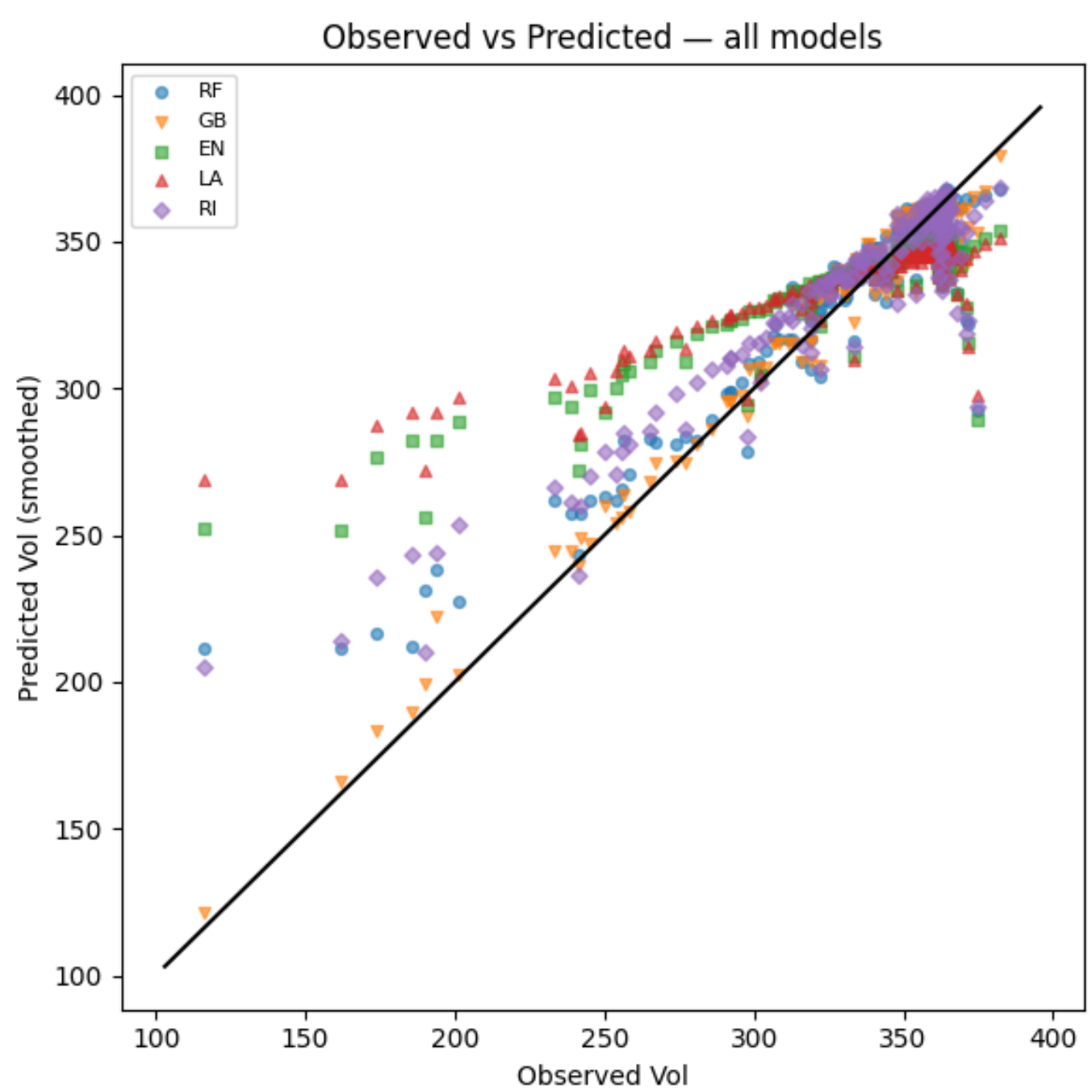

***Figure 8*** *Observed-versus-predicted scatter for all models (2000–2024). The 1:1 line is shown in black. Each point represents a daily storage estimate. Ridge (pink ◆) and RF (blue ●) cluster near the line, indicating unbiased performance. GB (orange ▲) shows a*

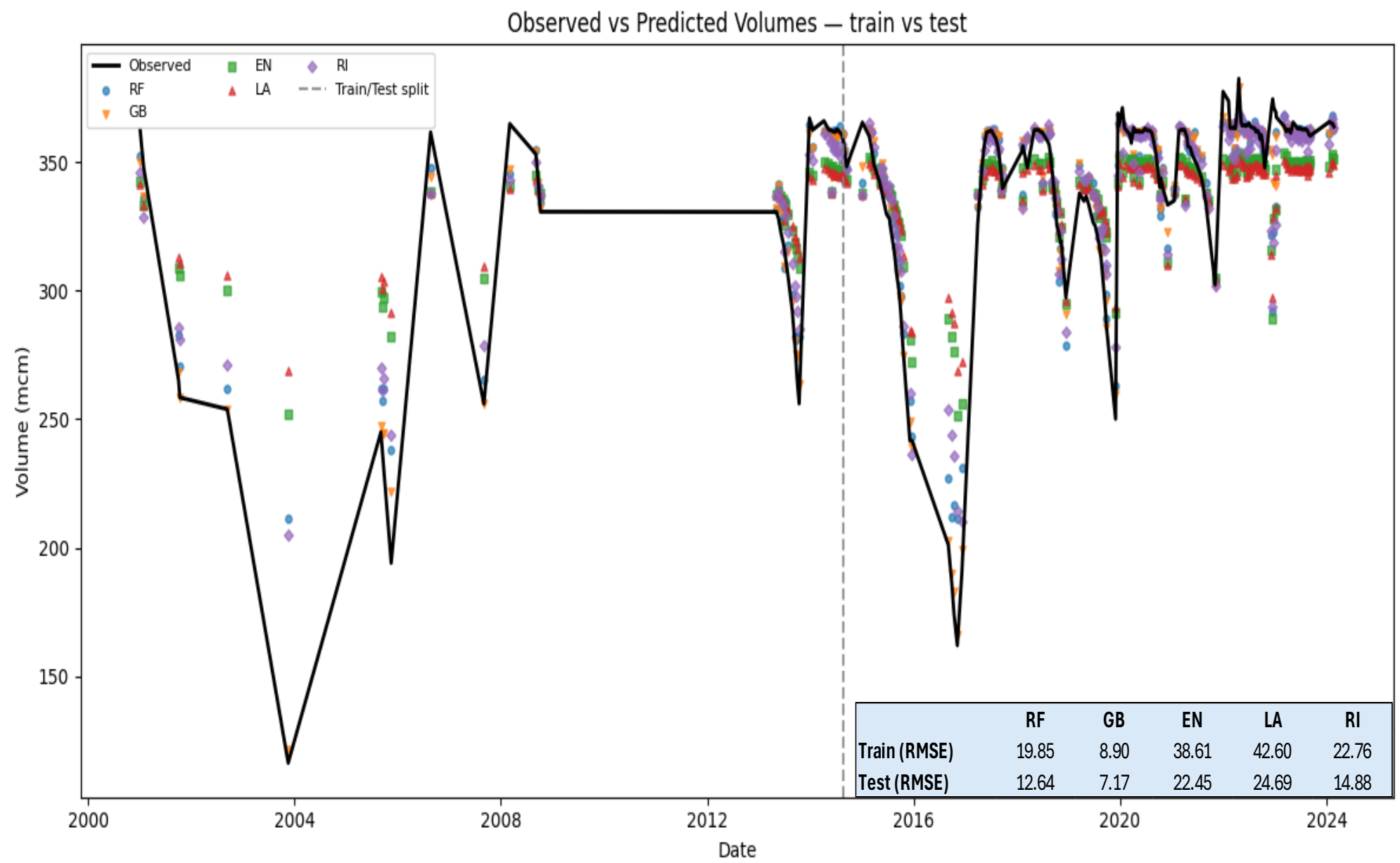

***Figure 9*** *Observed-versus-predicted timeseries for all models (test vs train) (2000–2024). The associated RMSE values for the test and train data of each model are shown in 'Train (RMSE)' and 'Test (RMSE)'.*

### 4.5 Operational implications

In normal storage years (≥ 250 MCM), Ridge, Forest, and Gradient Boosting all meet the operational accuracy target of ±5% of capacity. However, Ridge edges RF by ≈ 0.5 MCM and GB by ≈ 2 MCM in cross-validation (**Table 2**). In drought or maintenance draw-down periods (< 250 MCM), Ridge's advantage grows to ≈ 5 MCM over GB and ≈ 6 MCM over RF. The stacked ensemble inherits GB's high-band fidelity while retaining Ridge's robustness at low storage, providing a single model that satisfies both use cases (full-range RMSE ≈ 11 MCM, < 3% of capacity). Accordingly, we recommend the following deployment strategy for Loskop Dam.

- Ridge (Full-Capacity features) – for early-warning dashboards, where under-prediction of critically low volumes must be avoided. Ridge's conservative bias (tending to overestimate low volumes slightly) reduces the risk of declaring "water available" when the reservoir is depleted (Ford & Sankarasubramanian, 2023).
- Gradient Boosting (Full-Capacity, retrained daily) – for routine short-horizon forecasts during normal operations. GB provides the lowest in-sample error and can capture subtle non-linear curvature during the wet season. Frequent retraining ensures it adapts to the latest inflow patterns.
- Stacked ensemble (Ridge meta-learner) – for all-weather web services or dashboards that must operate across the full hydrological envelope without manual model switching. The stacker provides a balanced performance (CV RMSE ≈ 11 MCM) and gracefully handles transitions between regimes.

Additionally, Random Forest (with Full-Capacity features) is a viable fallback on edge devices or simpler implementations: its accuracy is within 0.6 MCM of Ridge in both bands (**Table 3**), and it requires no hyperparameter tuning.

The following steps are advisable to maintain optimum performance over time:

Quarterly model retraining using a rolling 5-year Timeseries Split, to update hyper-parameters as new data arrive.

- Regime-specific monitoring – compute Low/High band RMSE after any significant bathymetric change or rating-curve revision, to verify that accuracy remains within the ±5% target; and
- Transparent reporting – publish the global CV results and regime-specific error charts (similar to Table 2 and Table 3) in technical reports, enabling stakeholders to trace which regime informed each operational rule.

With these safeguards, the Loskop Dam prediction suite meets the < 5% RMSE criterion across its active storage range while maintaining transparency about limitations under extreme draw-down conditions.

## 5. Discussion

5.1 Why Gradient Boosting "looks" good yet underperforms out of sample

When all 24 years of data are supplied to the algorithm, Gradient Boosting (GB) learns a large ensemble of shallow trees that can mirror both the low- and high-volume regimes with very small residuals; hence its tight clustering around the 1:1 line in **Figure 8** and the in-sample RMSE of 6.8 MCM in the low band (**Table 4**). This behaviour is typical of boosting models, which reduce bias at the expense of variance (Friedman, 2001). In the 5-fold Timeseries Split evaluation, the same flexibility becomes a liability: each test fold contains hydro-climatic patterns (e.g., 2015–16 drought) absent from the training slice, and the over-fitted trees extrapolate poorly. As a result, GB's cross-validated RMSE balloons to ~25 MCM in the low regime (**Table 3**). Ridge regression, constrained by an $\ell^2$ penalty, exhibits higher bias but significantly lower variance; its learned coefficients change minimally from fold to fold, resulting in the best out-of-sample accuracy.

5.2 Regime-specific skill and operational risk

Loskop Dam typically operates between 250 MCM and 370 MCM, but historical lows of ~185 MCM were recorded in 2004 and 2016. In these drought years, GB and RF both exhibit a systematic negative bias (under-prediction of storage), whereas Ridge remains conservative, with an RMSE of 18.0 MCM compared to 22–25 MCM for the tree models (**Table 3**). A conservative bias is preferable for early warning because it reduces the chance of assuming "water available" when the reservoir is critically low (Ford & Sankarasubramanian, 2023). In the normal-storage range (≥ 250 MCM), the differences shrink: Ridge and RF tie at ≈ 12 MCM; GB is ~14–15 MCM—still within the 5% accuracy threshold set by South African water-authority guidelines (DWS 2021). The small margin confirms that once the surface area approaches its asymptote, the area–volume relationship is almost linear; a simple linear model suffices in the high regime, and the flexibility of boosting offers a slight advantage.

5.3 Value of feature inclusion

Adding the power-law calculated volume $\hat{V}$ Reduced global RMSE by ≈ 1 MCM for every algorithm, confirming prior work that reservoir morphometry is a primary driver of volume–area conversion (Gao et al., 2017). By contrast, the eight river-geometry variables ($R_3$ features) cut error by < 0.2 MCM on average, which aligns with remote-sensing studies that found limited incremental gain once the DEA surface-area series is available (Digital Earth Africa, 2023). Elevation ($R_4$) added virtually no new information because, in a single-pool reservoir, surface area at a given time is already implicitly linked to the current water level. In summary, the Full-Capacity feature ($R_2$) set (surface area + $\hat{V}$) Appears to capture the essential bathymetric relationship, and more complex descriptors yield diminishing returns.

5.4 Stacked blending as a pragmatic compromise

Stacking with a Ridge meta-learner assigns positive weights to GB in high-volume periods (where its bias is low) and to Ridge/RF during droughts. The resulting ensemble reduces the full-series RMSE to ≈11 MCM (~3% of live storage) without introducing any manual switches or non-linearities. This improvement echoes similar gains reported for reservoir-inflow prediction by Deb et al. (2024), who also found that blending multiple algorithms outperformed individual models in highly variable hydrological series. In our case, the stacker inherits GB's precision in tracking peak storage while compensating for GB's tendency to underpredict extreme lows (**Figure 5**). The interpretability of the Ridge blender is an added benefit—it provides insight into which base model is trusted under which conditions.

5.5 Recommendations for operational deployment

**Table 3** and the preceding results suggest a tiered approach to the operational use of the models: routine storage monitoring (when volumes are within the normal range) can rely on a daily-updated GB model, which offers the best instantaneous fit. During drought alerts or planned maintenance drawdowns, the Ridge model is more dependable due to its stable performance in low-volume scenarios. For comprehensive decision-support systems (e.g., an online dashboard that must function across seasons), the stacked ensemble is the preferred choice, as it automatically balances the bias–variance trade-off across regimes. An additional consideration is computational simplicity: if resources are limited (for instance, on an embedded system at the dam), the Random Forest model, though slightly less accurate, requires no parameter tuning and runs quickly, making it a robust fallback.

In implementing these models, it is important to maintain an adaptive and transparent framework. Regular re-calibration (every quarter or after major hydrological events) will update the model coefficients and ensure the volume–area relationship remains current despite sedimentation or other shifts. Moreover, communicating the model's accuracy to stakeholders is key. Publishing an updated performance table and error charts (as in **Table 1** and **Table 2**) in quarterly reports can build confidence in the system and highlight any conditions under which accuracy degrades. By following these practices, the Loskop Dam storage estimation models could potentially meet the acceptable error standard of the DWS, while providing actionable information for real-time water management decisions.

The IWMI Limpopo Digital Twin (LDT) Initiative, developed in close collaboration with partners and stakeholders, aims to improve water resource management and conservation in the Limpopo River Basin (LRB). As a foundational component, a daily time-step hydrological model of the basin was developed using the SWAT+ tool to address the absence of a publicly available water resources assessment framework. To enhance the model's capability for evaluating sub-basin water availability, dam volume prediction data were integrated into the SWAT+ model. Since the SWAT+ model's performance relies heavily on accurate dam volume data, ensuring the reliability of this data is critical (Gurusinghe et al., 2024; Garcia Andarcia et al., 2024)

## 6. Conclusion

This study provides a comprehensive evaluation of various machine learning models—Ridge Regression, Random Forest, Gradient Boosting, Lasso, and Elastic-Net—on their ability to predict water storage volumes in Loskop Dam. The results highlight that the choice of algorithm should be aligned with the specific operational needs of the prediction task, particularly in terms of the hydrological regime and forecast horizon.

Ridge Regression (RR) emerged as the most reliable model, especially under low-storage (drought) conditions, due to its stability and low variance, making it ideal for long-term predictions and early warning systems. The Ridge model's conservative bias ensures a lower risk of overestimating water availability during critical periods. For routine, short-term forecasts during normal operations, Gradient Boosting (GB) provides the best fit, benefiting from its ability to capture intricate non-linear relationships when retrained frequently. The stacked ensemble, combining the strengths of Ridge, RF, and GB, outperformed individual models by maintaining robust accuracy across both high and low storage regimes, without the need for manual switching. This ensemble model is well-suited for integrated, all-weather dashboards or web services that must operate across the full hydrological envelope.

The study further underscores the value of feature selection, where the "Full-Capacity" feature set, which combines water surface area and volume, provided the most significant improvements in predictive accuracy. Additional features such as geographical and elevation data showed only marginal benefits, confirming that the volume-area relationship is the primary driver of storage prediction.

In conclusion, the appropriate model selection depends on the specific use case and operational requirements: Ridge and the stacked ensemble models are best suited for long-term, cross-season forecasts, while Gradient Boosting excels in short-term predictions when retraining is feasible. This comprehensive assessment provides a robust framework for operational deployment, ensuring that the chosen model meets both the needs of real-time decision-making and the long-term accuracy goals, with a clear strategy for model maintenance and recalibration.

**Acknowledgement**

This publication was produced under the CGIAR Initiative on Digital Innovation, which advances sustainable agrifood systems through the development and application of digital solutions. We are thankful to the Leona M. and Harry B. Helmsley Charitable Trust, The Limpopo Watercourse Commission (LIMCOM), and all data and tool providers. The boundaries, names, and designations on maps do not imply endorsement by IWMI, CGIAR, partner institutions, or donors. We also wish to thank all funders who supported this research through their contributions to the CGIAR Trust Fund (https://www.cgiar.org/funders/).